\pdfoutput=1

\documentclass[11pt]{article}

\usepackage{ACL2023}
\usepackage{graphicx}
\usepackage{booktabs}
\usepackage{tabularx}

\usepackage{times}
\usepackage{latexsym}

\usepackage[T1]{fontenc}

\usepackage[utf8]{inputenc}

\usepackage{microtype}

\usepackage{inconsolata}

\usepackage{cleveref}
\usepackage{tikz}

\usepackage[normalem]{ulem} 

\usepackage{hyperref}
\usepackage[hidecomments]{ufdatastudio}
\usepackage{multirow}

\title{Retrieval-Augmented Long-Context Translation for Cultural Image Captioning: Gators submission for AmericasNLP 2026 shared task}

\author{
  Aashish Dhawan \\
  University of Florida \\
  \texttt{aashish.dhawan@ufl.edu}
  \And
  Christopher Driggers-Ellis \\
  University of Florida \\
  \texttt{driggersellis.cw@ufl.edu}
  \And
  Dzmitry Kasinets \\
  University of Florida \\
  \texttt{dkasinets@ufl.edu}
  \AND
  Christan Grant \\
  University of Florida \\
  \texttt{christan@ufl.edu}
  \And
  Daisy Wang \\
  University of Florida \\
  \texttt{daisyw@cise.ufl.edu}
}

\begin{document}
\maketitle

\newcommand{\grn}{Guaraní }
\newcommand{\bzd}{Bribri }
\newcommand{\yua}{Yucatec Maya }
\newcommand{\nlv}{Orizaba Nahuatl }
\newcommand{\hch}{Wixárika }


\begin{abstract}
    We present the University of Florida Gators submission to the AmericasNLP 2026 shared task on cultural image captioning for Indigenous languages. Our two-stage pipeline generates a Spanish intermediate caption with Qwen2.5-VL, then produces the target-language caption using retrieval-augmented many-shot prompting with Gemini 2.5 Flash. We achieve  164.1\%, 131.7\%, and 122.6\% improvements over the shared task baseline for Bribri, Guaraní, and Orizaba Nahuatl captioning, respectively,  in our dev set evaluation and maintain >150\% improvements for the Bribri and Orizaba Nahuatl languages in the test set evaluation. We find retrieval is highly language-dependent, beneficial only for large, in-domain corpora, and that synthetic data augmentation accounts for around 28 chrF++ of the dev set Guaraní performance gain.
    Our submission is the overall winner of the shared task, placing second out of five finalist submissions in human evaluations of target-language captions. Code and prompts are available on GitHub.\footnote{\url{https://github.com/dhawan98/AmericasNLP2026-Gators-Submission}}
\end{abstract}

\section{Introduction}\label{sec:introduction}
The AmericasNLP 2026~\cite{bui-etal-2026-findings} consists of generating culturally grounded captions for images in Indigenous languages of the Americas. We identify three major challenges. First, the target languages are low-resource. Second, the captions are culturally specific rather than generic visual descriptions. Third, the task requires not only lexical transfer, but also stylistic control.
Successful systems must produce short, natural captions that match the reference register expected by the organizers.


While the cultural image captioning task is new to AmericasNLP 2026, we find that current vision-language models (VLMs) are not able to directly caption images in the target languages.
This is unsurprising given the limited training data available for the target languages.
We therefore formulate the problem as a compound of image captioning in a high-resource language followed by machine translation from that language into one of the shared task targets, which mirrors the shared task's baseline approach.

We first attempt to build on the work of \citet{dhawan-etal-2026-improving} by providing intermediate Spanish (Es) VLM captions to an mBART-based machine translation (MT) model.
\citet{dhawan-etal-2026-improving} show that synthetic parallel data and language-specific preprocessing improve low-resource Indigenous MT, including Es-\grn (Grn) translation. 
In the 2026 shared task, however, a standard neural machine translation pipeline proves insufficient. When we apply the existing Es-Grn translation model to Spanish intermediate captions produced from the dev images, the resulting captions are often fluent enough at the sentence level but do not match the target caption register. 
In other words, the performance bottleneck shifts from generic translation quality to domain adaptation and culturally grounded caption style.

To address this mismatch, we move from sequence-to-sequence translation toward retrieval-augmented in-context translation with Large Language Models (LLMs). 
The core idea is simple. 
Instead of relying on a single model fine-tuned on mixed-domain low-resource language corpora, we retrieve Es-low-resource (LoRes) examples that are similar to the current caption and provide them as in-context exemplars. 
This design lets the decoder adapt its lexical choices and stylistic register at inference time. We evaluate multiple OpenAI and Gemini models, vary the number of retrieved training examples and development exemplars, and test prompt variants and retrieval heuristics.

Our experiments lead to three main findings. 
First, among OpenAI models, many-shot direct translation outperforms both our mBART baseline and LLM post-correction, but gains are modest and highly sensitive to prompt composition. 
Second, Gemini 2.5 Flash ~\cite{comanici2025gemini25} is dramatically stronger in Es-LoRes in-context translation than the GPT-family models. 
Third, development references are useful as in-context exemplars for matching the expected caption register, but they must be interpreted carefully because same-pool development prompting can inflate development-set scores.

Our primary contributions are as follows.
We present a retrieval-augmented LLM caption translation pipeline for low-resource cultural image captioning. 
We document an extensive set of negative and positive ablations, including model substitutions, prompt revisions, and development-exemplar hyperparameter searches. 
Finally, we highlight evaluation design as a central methodological issue for dev-conditioned in-context learning in low-resource captioning.

\section{Background}\label{sec:background_related_work}
Machine translation for Indigenous languages of the Americas has advanced largely through the AmericasNLP shared tasks, which have established evaluation benchmarks and made parallel corpora available for many low-resource languages \cite{mager2021findings, ebrahimi-etal-2023-findings, ebrahimi2024findings}. Strong systems in these shared tasks have typically relied on multilingual pretrained models such as mBART, M2M-100, or NLLB-200, often combined with synthetic data generation, multilingual transfer, and additional corpus collection \cite{gow2023sheffield, tonja2023enhancing, costa2022no}. A recurring pattern in this literature is that performance improvements come not only from larger models, but also from better alignment between model capacity, augmentation strategy, and the target domain.

Character-based evaluation metrics are also especially relevant in this area. chrF and chrF++ are widely used for morphologically rich languages because they are more robust than BLEU to inflectional variation and spelling differences \cite{popovic-2015-chrf, popovic-2017-chrf}. The AmericasNLP 2026 organizers likewise use chrF++ as the first-stage ranking metric for the shared task.

Finally, our work relates to the broader use of in-context learning for low-resource generation. Instead of relying exclusively on fixed model parameters after fine-tuning, retrieval-augmented prompting can adapt the decoder to the current example at inference time. In our setting, this is especially attractive because the target outputs are short and stylistically constrained: parallel demonstrations can serve not only as semantic guides, but also as direct evidence of the desired caption register.

\section{Dataset and Methodology}\label{sec:datsets_methodology}

\subsection{Baseline System}

We compare our retrieval-augmented LLM pipeline against the Qwen3VL-8B \cite{qwen3technicalreport} Captioning and Sheffield 2023 \cite{gow2023sheffield} MT method, which is the stated baseline for the current shared task on image captioning. 
We apply this baseline to our submission results for dev set captioning in each of the applicable target languages and report percent improvement over the baseline scores.

For MT in \grn captioning, we also compare against the mBART-based Es-Grn model from \citet{dhawan-etal-2026-improving}, trained on curated and synthetic parallel data. This serves as the strongest conventional translation baseline in our pipeline and allows us to assess whether retrieval-augmented in-context generation improves over standard sequence-to-sequence translation.
We evaluate several retrieval-augmented GPT-4-family~\citep{openai2024gpt4technicalreport} model variants, summarized in Table~\ref{tab:mt_ablation}: GPT-4o-mini, GPT-4.1-mini, GPT-4.1, along with the competing Gemini 2.5 Flash. 

Across these models, we vary the number of retrieved training examples $r$, development exemplars $d$, and prompting strategy.

\subsection{Task Data}

The official development set $\mathcal{D}$ (dev set) contains 50 examples, and the organizers release the data in JSONL format paired with images. The pilot set includes Spanish captions for reference, but the organizers explicitly note that these are pilot-only and will not be present in development or test.

Our final submission pipeline is two-stage. Stage 1 produces a Spanish caption from the image using Qwen 2.5B \cite{bai2025qwen25vl}. Stage 2 utilizes Gemini 2.5 Flash \cite{comanici2025gemini25} for Es-LoRes translation.







\subsection{Datasets by Language}

The shared task covers five target languages: Guaraní, Yucatec Maya, Orizaba Nahuatl, Bribri, and Wixárika. As shown in \Cref{tab:dev_results}, the available retrieval data $\mathcal{R}$ differs substantially across languages, which motivates treating the retrieval size $r$ and the number of development exemplars $d$ as inference-time hyperparameters that vary across target language submissions.

\paragraph{\grn} We use the largest retrieval bank in our setup. The retrieval Es-Grn bank of 53,183 pairs contains  AmericasNLP 2023 \cite{ebrahimi-etal-2023-findings} training data augmented with synthetic examples from the MultiScript30k project \cite{driggersellis2025multiscript30kleveragingmultilingualembeddings}. 
The \grn retrieval dataset is relatively well aligned with the captioning task, as it contains culturally specific terms, proper nouns, and short descriptive examples useful for visual caption generation.

\paragraph{\yua} We do not have a comparable parallel training corpus for retrieval. The development set $\mathcal{D}$ is the only retrieval source, so we rely on dev exemplars and the pretrained knowledge of the LLM and we fix $r = 0$ in all experiments.

\paragraph{Other Target Languages} We use the available Es-LoRes parallel data as retrieval banks. Their retrieval banks vary in size and domain match: \nlv has 16,145 pairs, \bzd has 7,508 pairs, and \hch has 8,966 pairs. The \hch retrieval data is notably less caption-like, since much of the available corpus is narrative or literary rather than visual-description oriented. These differences motivate the language-specific hyperparameter choices reported in \Cref{tab:dev_results}.

\subsection{Retrieval Bank and Prompt Construction}

For each target language, we construct a language-specific Es-LoRes retrieval bank $\mathcal{R}$ from the available parallel data described above. The Spanish side of each retrieval bank is indexed with BM25 \cite{robertson2009probabilistic}, a TF-IDF-style lexical retrieval method that ranks candidate examples by query-term overlap while accounting for term importance and document length normalization. At inference time, the Spanish caption generated in Stage~1 is used as a query $q$, and the top-$r$ retrieved pairs are selected as $\mathcal{R}_r(q) \subset \mathcal{R}$.

In addition to retrieved training pairs, some configurations include development exemplars. Let $\mathcal{D}$ denote the shared-task development set for a target language, and let $\mathcal{D}_d \subset \mathcal{D}$ denote the $d$ development examples included in the prompt. For a query caption $q$, the full prompt context is defined as
\[
P(q) = \mathcal{D}_d \cup \mathcal{R}_r(q),
\]
where $\mathcal{R}_r(q)$ provides retrieval-based semantic and lexical grounding, and $\mathcal{D}_d$ provides examples of the caption style expected by the task. The model then generates the target-language caption conditioned on $P(q)$.

We sweep $r$ and $d$ where applicable. The values selected for submissions are in  \Cref{tab:dev_results}, and detailed grid-search results appear in our appendix.

\subsection{Prompting Strategy}

The MT prompt structure is deliberately simple. The system prompt instructs the model to translate from Spanish into the target language, match the example style, stay concise, preserve culturally specific nouns when appropriate, and produce exactly one line. The user prompt contains two evidence blocks, development exemplars $\mathcal{D}_d$ and retrieved Spanish--target-language pairs $\mathcal{R}_r(q)$, followed by the current Spanish caption.

We use the same general prompt structure across target languages, adding language-specific modifications only when required. After several prompt-engineering attempts, we found that more aggressive instructions, such as suppressing generic lead-ins or forcing noun-first phrasing, reduced performance. The final prompts therefore keep the system instruction minimal and rely on in-context examples for lexical and stylistic guidance. We summarize the final and ablation prompt files in \Cref{tab:prompts} of the appendix.

\subsection{University of Florida Gators Submission}

\begin{figure}[!htbp]
    \centering
    \includegraphics[width=\columnwidth]{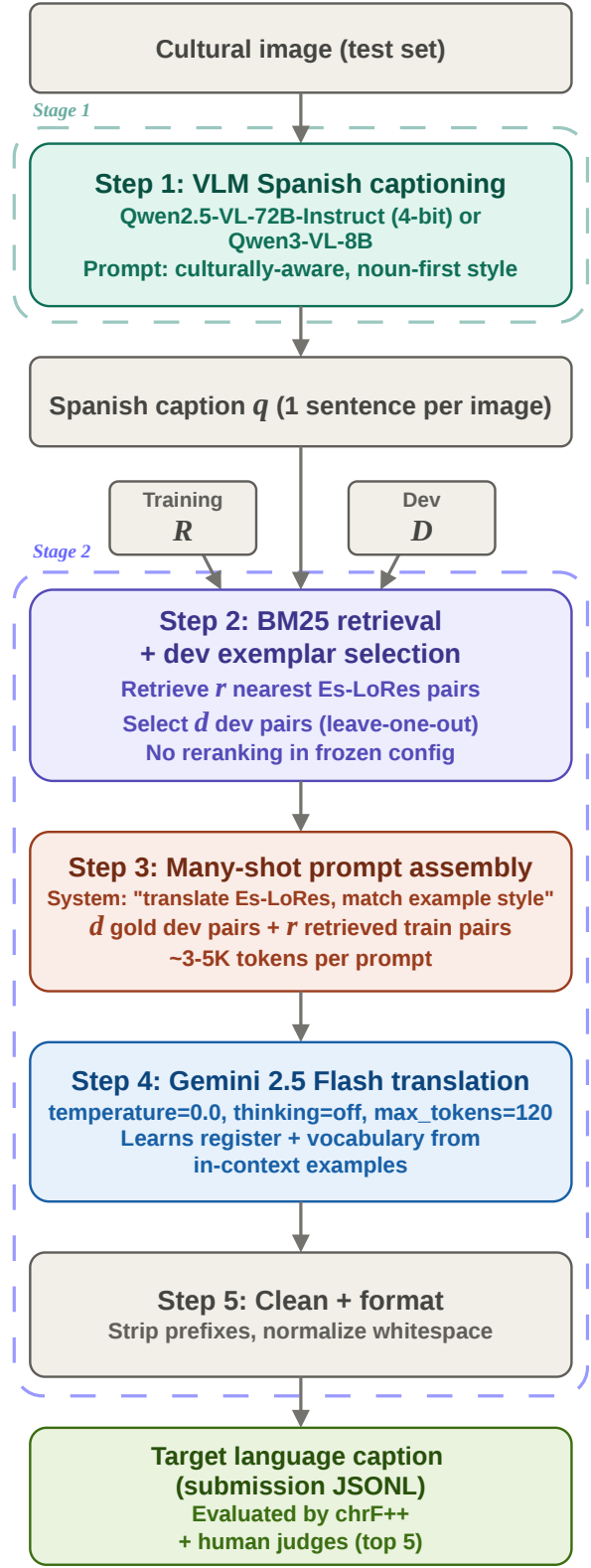}
    \caption{Overview of the proposed two-stage image captioning pipeline. Here, $\mathcal{R}$ denotes the full Es-LoRes retrieval bank, $\mathcal{R}_r(q) \subset \mathcal{R}$ denotes the $r$ nearest Es-LoRes training pairs retrieved for Spanish caption $q$, $\mathcal{D}$ denotes the full development set, and $\mathcal{D}_d \subset \mathcal{D}$ denotes the $d$ gold dev pairs used in the many-shot prompt.}
    \label{fig:pipeline}
\end{figure}

Our submission system features a retrieval-augmented long-context translator embedded in a two-stage image captioning pipeline. Figure~\ref{fig:pipeline} illustrates the full system, which is organized into five steps. Stage~1 corresponds to Step~1 in the diagram. A vision-language model generates one Spanish caption $q$ for each target image. We use either Qwen2.5-VL-72B-Instruct in 4-bit precision or Qwen3-VL-8B~\cite{bai2025qwen3vl} for this stage. The prompt is culturally aware; follows a noun-first style; and encourages concise descriptions of visible entities, objects, clothing, actions, and scene context. We treat this stage as fixed and do not optimize it extensively in this paper.

Stage~2 corresponds to Steps~2--5 in Figure~\ref{fig:pipeline}. It transforms the generated Spanish caption $q$ into the final target-language caption using retrieval-augmented many-shot MT. The $q$ is used as a BM25 query over the Spanish side of the available Es-LoRes retrieval bank $\mathcal{R}$. In Step~3, the retrieved subset $\mathcal{R}_r(q)$ and development subset $\mathcal{D}_d$ are assembled into a many-shot prompt with an instruction to translate from Spanish to the target language while matching the example style. Each prompt contains approximately 3K--5K tokens. The $r$ nearest Es-LoRes training pairs provide semantic and lexical grounding, while the $d$ development pairs provide direct evidence of the target caption register. In Step~4 of Figure~\ref{fig:pipeline}, Gemini 2.5 Flash performs the final target-language generation with temperature set to 0.0, thinking disabled, and a maximum output length of 120 tokens. We then strip prefixes in Step~5 and normalize whitespace to produce the final JSONL submission. The system is evaluated using chrF++, followed by human judgment for the top-ranked submissions.

\begin{table*}[t]
\centering
\small
\begin{tabular}{lrrrr|rr|rr}
\toprule
\multicolumn{5}{c}{\textbf{Configuration}} & \multicolumn{2}{|c|}{\texttt{test}} & \multicolumn{2}{c}{\texttt{dev}} \\
\midrule
Language & \multicolumn{1}{r}{$r$} & \multicolumn{1}{r}{$d$} & \multicolumn{1}{l}{Test Examples} & \multicolumn{1}{l}{Retrieval Pairs ($R$)} & \multicolumn{1}{|r} {Test chrF++} & \multicolumn{1}{r|}{\% Imp.} & \multicolumn{1}{r} {Dev chrF++} & \multicolumn{1}{r}{\% Imp.}\\
\midrule
\bzd & 80 & 20 & 267 & 7,508 & 17.90 & \underline{155.2\%} & 19.99  & \textbf{164.1\%} \\
\grn & 80 & 49 & 101 & 53,183* & \underline{23.10} & 14.72\% & \textbf{48.24} &  \underline{131.7\%} \\
\nlv & 40 & 20 & 200 & 16,145 & \textbf{25.42} & \textbf{166.9\%} & 25.67 & 122.6\% \\
\hch & 40 & 20 & 201 & 8,966 & 17.58 & 3.952\% & 18.99 & 6.866\% \\
\yua & -- & 49 & 212 & 0$\dagger$ & 21.11 & -- & \underline{26.29} & --\\
\bottomrule
\end{tabular}
\caption{Submission chrF++ results across languages for the proposed system. Here, $r$ is the number of retrieved training pairs included in each prompt, $d$ is the number of development exemplars included in each prompt, and $R$ denotes the total size of the available retrieval bank for each language. Baseline results are those provided for the shared task \cite{gow2023sheffield}. \textbf{Bold} and \underline{underlined} entries in the right-side columns indicate the best and second-best results, respectively. * Includes parallel Es-Grn MultiScript30k \cite{driggersellis2025multiscript30kleveragingmultilingualembeddings} synthetic exemplars. $\dagger$ No external retrieval bank is used; only development exemplars are included.}
\label{tab:dev_results}
\end{table*}

\section{Results}\label{sec:results}

\begin{table*}[t]
\centering
\small
\begin{tabular}{lrrp{4cm}p{4cm}r}
\toprule
System & \multicolumn{1}{r}{$r$} & \multicolumn{1}{r}{$d$} & Notes & \multicolumn{1}{c}{Setting type} & \multicolumn{1}{r}{chrF++} \\
\midrule
mBART Baseline              & -- & -- & Inherited MT Baseline        & Baseline          & 22.40 \\
GPT-4o-mini Post-Correction & -- &  8 & Revise mBART Draft           & Post-Editing       & 19.81 \\
GPT-4o-mini                 & 24 &  4 & Direct Many-Shot Translation & Direct Prompting   & 23.41 \\
GPT-4o-mini                 & 28 &  4 & Best OpenAI Setting          & Direct Prompting   & 23.60 \\
GPT-4.1-mini                & 28 &  4 & Model Swap                   & Direct Prompting   & 23.23 \\
GPT-4.1                     & 28 &  4 & Model Swap                   & Direct Prompting   & 23.49 \\
Gemini 2.5 Flash            & 28 &  0 & No dev Exemplars             & Train-Only Prompt  & 32.14 \\
Gemini 2.5 Flash            & 28 &  4 & Same-Pool dev Prompting      & Dev-Assisted       & 39.25 \\
Gemini 2.5 Flash            & 28 & 20 & Same-Pool dev Prompting      & Dev-Assisted       & \underline{42.90} \\
Gemini 2.5 Flash            & 28 & 49 & Same-Pool dev Prompting      & Dev-Assisted       & 42.19 \\
Gemini 2.5 Flash            & 80 & 49 & Same-Pool dev Prompting      & Dev-Assisted       & \textbf{48.24} \\
\bottomrule
\end{tabular}
\caption{Dev chrF++ \grn captioning results across baseline machine translation and RAG-based prompting configurations. \textbf{Bold} indicates the best result, and \underline{Underline} indicates the second-best result.}
\label{tab:mt_ablation}
\end{table*}

\Cref{tab:dev_results} gives our final submission's performance for each of the target languages in the shared task while \Cref{tab:mt_ablation} summarizes the main progression of Es-Grn dev set experiments from an mBART translation model to in-context MT with Gemini 2.5 Flash, which is competent in all target languages.
Several trends emerge immediately. First, direct many-shot translation is better than post-correction, confirming that it is more effective to generate the caption in one step than to repair the mBART output after the fact. Second, among the OpenAI models we tested, GPT-4o-mini remains the strongest, but the gains over the mBART baseline are modest. Third, Gemini 2.5 Flash yields much larger improvements, even before adding any development exemplars.

In our final dev set results, the \grn target achieves the highest absolute chrF++ score among the five target languages by at least 10 chrF++ and more than doubles the \bzd and \hch target performances.
For \grn we achieve a remarkable 131.7\% improvement over the baseline method \cite{gow2023sheffield}.
In \Cref{sec:ablations}, we investigate the effect that synthetic exemplars may have had on \grn target language performance versus the other languages for which no MultiScript30k data exists.
However, as the final column in \Cref{tab:dev_results} attests, each language target outperforms the baseline method whenever the baseline is available.
In particular, for the \bzd and \nlv targets, respectively, we achieve 164.1\% and 122.6\% improvements over the baseline.
As we highlight in \Cref{tab:dev_results}, our improvement for \bzd is the greatest relative improvements over the baseline method for any target language.

The test results are slightly different. \bzd is replaced by \nlv as most improved language. The languages achieve 155.2\% and 166.9\% improvement over the testing baseline, respectively \cite{bui-etal-2026-findings}.
Guaraní performance drops by more than half in the test evaluation and most of the dev set performance gain is not reproduced in the official evaluation as a result. \hch performance improvement falls by a similar amount.

\section{Ablations}\label{sec:ablations}

In addition to the results we list in the previous section, we provide a number of ablations to demonstrate the superiority of the final submission over numerous alternative approaches to various facets of the captioning pipeline. 
In the experiments reported here, we hold the Spanish captions fixed and optimize only the \grn translation stage. 
This lets us analyze ablation effects independent of the image captioner and makes the ablations directly comparable to our dev set results.
We focus on the dev set evaluation here because ablations are all performed on the dev set before the release of final shared task results.

\subsection{Machine Translation Architecture}

We allude in the first section to how the initial approach for the machine translation step in \grn utilizes an mBART based MT model \cite{dhawan-etal-2026-improving}.
\Cref{tab:mt_ablation} shows Guaraní captioning performance across several MT architectures, including mBART, GPT-4o, and Gemini 2.5 Flash.
Where appropriate, we adopt the retrieval-augmented approach in our final submission, and we vary the values of $r$ and $d$ within architectures.

The results show that the best configuration for \grn translation is the Gemini 2.5 Flash LLM prompted with $r = 80$ and $d = 49$ exemplars.
We traverse much of the $r,d$ search grid with OpenAI models. The best OpenAI setting used GPT-4o-mini with 28 retrieved examples and 4 development exemplars, reaching 23.60 chrF++. This performance is reflected in \Cref{tab:mt_ablation}. 
Removing or increasing development exemplars reduced performance. 
GPT-4.1-mini and GPT-4.1 were both competitive but did not surpass GPT-4o-mini.
This pattern suggests that, for the OpenAI models we tested, the in-context regime has a narrow optimum. Too little context leaves the model underconstrained; too much context appears to add noise or dilute the style signal.

For its superior performance over GPT models and its positive receptivity to context, we adopt Gemini 2.5 Flash for Es-LoRes MT in our final submission.

\subsection{Hyperparameter Search}

For each language, we sweep values of $r$ and $d$ that we list in \Cref{sec:datsets_methodology}.
\Cref{tab:dev_results} of final submission results shows performance at the submittal configuration featuring its $r,d$ pair.
To be thorough, \Cref{tab:grid_search} reports the results of a partial grid search for each target language in our appendix.

Though this ablation shows that the $r$ and $d$ exemplar counts from \Cref{tab:dev_results} are optimal within our search grid, there is no accounting for values outside of it.
Additionally, performance changes as one scans the grid indicate differing impacts of the $r$ and $d$ hyperparameters for different languages and data sources.
Thus, we stress in this ablation the importance of a thorough search for the optimal $r,d$ pair for any new target language or data configuration in retrieval-augmented Es-LoRes MT.

\subsection{Synthetic Exemplars} \label{subsec:synthetic_exemplars}

We notice in \Cref{sec:results} that our pipeline performs \grn image captioning much more effectively in absolute terms versus any of the other target languages, even though we achieve similar percent improvement over the applicable baseline for the \bzd and \nlv targets.
We also note that \grn is the only target language for which we include synthetic exemplars from MultiScript30k.
Testing for the effect of this synthetic data, we ablate the original \grn captioning submission by only using AmericasNLP 2023 \cite{ebrahimi-etal-2023-findings} training data for retrieval exemplars ($r$).
\Cref{tab:ms30k_ablation} compares performance with and without MultiScript30k \cite{driggersellis2025multiscript30kleveragingmultilingualembeddings} synthetic exemplars for three $r, d$ pairs and provides the greatest performance overall without synthetic exemplars.

The results are clear.
Controlling for our retrieval hyperparameters by fixing $r$ and $d$ to three pairs, we observe that the original configuration with both genuine AmericasNLP 2023 training data and synthetic MultiScript30k exemplars outperforms the ablation with AmericasNLP 2023 alone by more than 100\% relative improvement in chrF++ in each case.
Additionally, the best \grn captioning performance without synthetic exemplars is 55.9\% less than the best performance with them.
These results mirror the comparison of our submission's \grn performance to the other target languages in the shared task and attribute much of our improvement in \grn captioning to synthetic exemplars in the retrieval-augmented Es-Grn translation step.

\begin{table}[t]
\centering
\small
\begin{tabular}{lrrrrr}
\toprule
Data & \multicolumn{1}{r}{$r$} & \multicolumn{1}{r}{$d$} & \multicolumn{1}{l}{Ret. Pairs} & \multicolumn{1}{r}{chrF++} \\
\midrule
ANLP2023 + MS30k & 40 & 49 & 53,183 & 51.34 \\
ANLP2023 + MS30k & 40 & 20 & 53,183 & 48.38 \\
ANLP2023 + MS30k & 80 & 49 & 53,183 & 48.24 \\
\midrule
ANLP2023 & 80 & 0  & 26,032 & 22.65 \\
ANLP2023 & 40 & 49 & 26,032 & 21.03 \\
ANLP2023 & 40 & 20 & 26,032 & 20.50 \\
ANLP2023 & 80 & 49 & 26,032 & 20.75 \\

\bottomrule
\end{tabular}
\caption{Dev chrF++ results for \grn captioning with differing retrieval exemplar sets. Data includes AmericasNLP2023 (ANLP2023) training data \cite{ebrahimi-etal-2023-findings} and/or MultiScript30k (MS30k) \cite{driggersellis2025multiscript30kleveragingmultilingualembeddings} synthetic exemplars as noted in the column \textit{Data}.}
\label{tab:ms30k_ablation}
\end{table}

\subsection{Alternative Prompting and Reranking}
We also test several prompting ablations that looked reasonable but are consistently negative in their impact on performance, with the exception of a specific strategy we give additional attention to in \Cref{subsec:morph_bzd}.
In this section, we quickly summarize other alternative prompts, and in \Cref{tab:prompts}, we produce all of the relevant prompts for completeness.

A prompt rewrite aimed at suppressing generic scene-introduction phrases reduces performance substantially, and a retrieval reranker that attempts to prefer short caption-like pairs also reduces performance, both for OpenAI and Gemini. 
Likewise, using 49 same-pool development exemplars with GPT-4o-mini degraded performance instead of improving it.
These results matter because they show that this task does not respond well to aggressive heuristic control. 
The most effective systems are built by keeping the instruction stable and varying only model choice and amount of context supplied.

For \hch in particular, we devise special prompts with cultural context in the form of a glossary.
These glossaries contain the names of common objects from the \hch culture and their definitions in Spanish, with the hope that this additional context will help the LLM MT architectures translate intermediate Spanish captions into the \hch target language.
We utilize two versions of the cultural glossary, but the results do not improve for either one.
We include the additional prompts in \Cref{tab:prompts} to document this alternative prompting.

\subsection{Morphological Considerations for \bzd} \label{subsec:morph_bzd}

We ablate our submission for the \bzd (Bzd) language by considering performance with and without morphological prompting and post-processing for the Es-Bzd MT.
\bzd has complex tonal marking in orthography (circumflexes, underlines, multiple diacritics) which makes the deduplication logic harder \cite{coto-solano-2021-explicit}.
The tokenizer splits differently on tonal characters.

In our final submission, we utilize a complex postprocessing and additional prompting for the Gemini MT model to improve performance. 
\Cref{tab:morph_ablation} shows the evolution of our strategies to account for the morphological complexity of the \bzd language.
First, careful examination of our pipeline's output reveals that \bzd captions are initially in NFC encoding, which combines base letter and diacritic encodings into one character.
This does not match the dev set examples, which use NFD encoding.
For instance, our model outputs \textit{ë} as a single unit, but the dev set's NFD encoding would separate the letter \textit{e} from the umlaut above it.
Because chrF++ is a character-level lexical metric, this mismatch depresses the score. We therefore perform NFD-Normalization (NFD-Norm.) to account for the difference.
Secondly, we believe the morphological considerations of Es-Bzd translation significant enough that they deserve special prompting.
To this end, we formulate a Morphological Prompt (Morph.) for \bzd that includes special instructions about Subject-Object-Verb (SOV) word order, tonal diacritics, common consonant clusters, verb-final clauses, and possessive noun prefixes.

As in previous ablations, the results in \Cref{tab:morph_ablation} largely speak for themselves. NFD-normalization overcomes the encoding mismatch, and morphological prompting provides additional improvements.

\begin{table}[t]
\centering
\small
\begin{tabular}{lrrr}
\toprule
Language & Prompting & Postprocessing & \multicolumn{1}{r}{chrF++} \\
\midrule
\bzd & Standard & Standard & 11.50 \\
\bzd & Standard & NFD-Norm. & 17.02 \\
\bzd & Morph. & NFD-Norm. & 19.99 \\
\bottomrule
\end{tabular}
\caption{Morphological ablations for our submission in the \bzd captioning task. We either utilize NFD-Normalization (NFD-Norm.) as a postprocessing step, both NFD-Norm. and Morphological Prompting (Morph.), or we apply the same prompting and postprocessing as other target languages (Standard). Rightmost column shows Dev ChrF++.}
\label{tab:morph_ablation}
\end{table}

\section{Discussion}\label{sec:discussion}

We now proceed to a discussion of our method and the results achieved.
We analyze the available data and investigate the potential sources of performance improvement over the baseline method.

\subsection{In-Context Retrieval}
Though limited to the \grn target, we show in our ablation of MT architectures that long-context retrieval-based MT using state-of-the-art LLMs greatly improves image captioning performance at the MT stage versus the dedicated mBART model \cite{dhawan-etal-2026-improving}.
While we cannot say for certain whether this relationship holds for other language targets, from comparisons to baseline performances, it appears that additional context from real and synthetic exemplars significantly boosts LLM translation in Es-LoRes tasks. 

\subsection{Synthetic Exemplars}

For the \grn language, we include approximately 30k synthetic exemplars for in-context retrieval from the MultiScript30k \cite{driggersellis2025multiscript30kleveragingmultilingualembeddings} dataset.
We observe in Section 4 that Guaraní achieves substantially higher absolute dev chrF++ than the other target languages.

Remembering that synthetic exemplars apply only to \grn and its significantly positive effect on Es-Grn MT in related work \cite{dhawan-etal-2026-improving}, we ablate for the synthetic exemplars' effect on \grn captioning.

For \grn captioning without the MultiScript30k synthetic retrieval pairs, the results we elaborate in \Cref{subsec:synthetic_exemplars} and \Cref{tab:ms30k_ablation} show a comparison similar to \grn captioning versus the other target languages.
Results improve over 100\%, approximately 28 total chrF++ or more, for \grn captioning with the synthetic exemplars for the same $r,d$ pair.
These results also mirror observations from \cite{dhawan-etal-2026-improving} on the effect of synthetic exemplars on Es-Grn MT.
We therefore conclude that synthetic exemplars are a driver of \grn performance and speculate that synthetic retrieval pairs may further improve captioning performance for other target languages.

\subsection{Morphology}
Despite favorable results in the \grn target, we acknowledge in \Cref{sec:results} that our greatest relative improvement is for the \bzd language.
The ablation in \Cref{subsec:morph_bzd} clarifies that much of this performance improvement stems from the morphological considerations we take for the \bzd language target via specialized prompting.
We observe that explicitly prompting for Bribri's morphological features accounts for over 10\% of the performance gain in our final submission over the shared task's baseline \cite{gow2023sheffield}.
We conclude that morphological prompting has potential for application in other low-resource, morphologically complex languages.

\section{Future Work}\label{sec:future_work}
The most impactful next step is improving the visual captioning stage. 
Error analysis indicates that approximately 54\% of remaining Guaraní errors originate in the vision model rather than the translator. 
A stronger VLM will likely yield larger gains than further translation tuning. 
Beyond captioning, we may extend retrieval to incorporate visual similarity rather than Spanish text overlap alone, which would help when the intermediate caption is noisy or culturally ambiguous. 
The lowest-resource languages (\hch and \bzd languages) are bottle-necked by corpus domain mismatch rather than model choice. 
Even small caption-style parallel data corpora for these languages would likely produce larger gains than any prompting improvement.

\section{Conclusion}\label{sec:conclusion}
We present the University of Florida Gators team system for the AmericasNLP 2026 shared task on cultural image captioning for Indigenous languages \cite{bui-etal-2026-findings}. 
Our two-stage pipeline, VLM captioning in Spanish followed by retrieval-augmented many-shot translation with Gemini 2.5 Flash, substantially outperforms fine-tuned baseline models across all five target languages, culminating at 48.24 chrF++ for Guaraní in the submission.
We find that retrieval behavior is highly language-dependent. 
Large retrieval windows help Guaraní but hurt Yucatec Maya, where Gemini's pre-training knowledge is sufficient and BM25 retrieval adds noise. 
We also find that development exemplars are useful for matching caption register, but their use requires careful interpretation because they can inflate development-set scores when drawn from the same evaluation pool.
For the lowest-resource languages, the ceiling is domain mismatch, not model capacity.

Finally, we hope our submission to the AmericasNLP 2026 shared task on cultural image captioning and our ablative analyses will guide future efforts in LoRes image captioning.

\section*{Limitations}\label{sec:limitations}
Our system is a cascade: errors in the Spanish captions propagate into translation with no recovery mechanism. Because most ablations hold the Spanish captions fixed and vary only the translation stage, they likely underestimate the contribution of the visual captioning model to final performance.

A second limitation is the use of development examples as in-context exemplars. These examples are useful for matching the expected caption register, especially when little caption-style target-language data is available, but they can also inflate development-set scores when exemplar selection and evaluation draw from the same small pool. We therefore treat dev-assisted results primarily as model-selection and submission-configuration evidence rather than as a fully independent estimate of generalization. A more rigorous evaluation would report held-out or cross-split dev results for each language.

Finally, evaluation relies primarily on chrF++, which is useful for low-resource and morphologically rich languages but cannot fully capture fluency, cultural appropriateness, or naturalness for native speakers. Although the shared task includes human evaluation for finalist systems, our own ablations do not include additional native-speaker evaluation.


\bibliography{citations}
\bibliographystyle{acl_natbib}

\newpage
\section*{Appendix} \label{sec:appendix}

Here, we provide additional data for validation of our method versus various ablations with particular focus on alternative prompts for various target languages and the $r,d$ grid search.
In the following sections, we give tables of alternative prompts and performance at different $r,d$ pairs for each target language.

\subsection*{Alternative Prompts} \label{subsec:alternative_prompts}

\Cref{tab:prompts} yields the prompts that we utilize in our ablations and the prompts present in
our final submission for each target language as indicated in the \textit{Standing} column. 
The \textit{Language(s)} column shows which languages the prompt applies to.

\subsection*{$r,d$ Hyperparameter Search}

We frequently refer to a search for optimal $r$ and $d$ retrieval hyperparameters in the main body of this paper but reserve detailed communication of the sweep for this appendix due to the number of configurations we consider.
For the Gemini 2.5 Flash MT architecture and for each target language, we partially sweep a grid consisting of $r,d$ pairs such that $r \in \{ 0, 10, 20, 40, 80 \}$ and $d \in \{0, 10, 20, 30, 40, 49\}$. Table~\ref{tab:grid_search} reports the chrF++ scores for the final captioning pipeline for each combination of $r$ and $d$ tested for each target language.

\begin{table*}[t]
\centering
\small
\setlength{\tabcolsep}{4pt}
\begin{tabular}{l | p{4cm} | l | c c}
\hline
Short Desc. & Description &  Prompt URL(s) & Standing & Language(s) \\
\midrule
Captioning & Prompts for VLM in first stage of captioning pipeline. &  \href{https://github.com/dhawan98/AmericasNLP2026-Gators-Submission/blob/main/prompts/v1_submission/guarani_caption_prompt.txt}{v1\_submission/guarani\_caption\_prompt.txt} & \textbf{Final} & All \\
\midrule
Many-Shot & Includes $r$ and $d$ in-context exemplars from previous parallel text corpora and from the dev set, respectively. & \href{https://github.com/dhawan98/AmericasNLP2026-Gators-Submission/tree/main/prompts/v1_submission}{v1\_submission/<target>\_system\_prompt.txt}  & \textbf{Final} & All \\
\midrule
Morph. \bzd & Accounts for \bzd target's morphological complexity. Portion shown appended to general prompt. & \href{https://github.com/dhawan98/AmericasNLP2026-Gators-Submission/blob/main/prompts/v1_submission/bribri_system_prompt.txt}{v1\_submission/bribri\_system\_prompt.txt} & \textbf{Final} & \bzd \\
\midrule
\hch Gloss. \textit{v2} & Provides definitions for culturally situated nouns in the \hch language. & \href{https://github.com/dhawan98/AmericasNLP2026-Gators-Submission/blob/main/prompts/wixarika/caption_es_wixarika_v2.txt}{wixarika/caption\_es\_wixarika\_v2.txt} & Ablation & \hch \\
\midrule
\hch Gloss. \textit{v3} & Provides definitions for culturally situated nouns in the \hch language. & \href{https://github.com/dhawan98/AmericasNLP2026-Gators-Submission/blob/main/prompts/wixarika/caption_es_wixarika_v3.txt}{wixarika/caption\_es\_wixarika\_v3.txt} & Ablation & \hch \\
\midrule
\end{tabular}
\caption{Prompts we utilize in our ablations and final submission for each of the target languages. The \textit{Standing} column indicates whether a prompt is an Ablation or part of our \textbf{Final} submission. The final \textit{Language(s)} column indicates what languages the prompt applies to.}
\label{tab:prompts}
\end{table*}
\begin{table*}[t]
\centering
\small
\begin{tabular}{lrr|p{6cm}|r}
\toprule

\multicolumn{5}{c}{\textbf{\textit{\bzd}}} \\
\midrule
Language & \multicolumn{1}{c}{Retrieval (r)} & \multicolumn{1}{r}{Dev Exemplars (d)} & \multicolumn{1}{|l|}{Notes} & \multicolumn{1}{r}{chrF++} \\
\midrule
\bzd & 80 & 20 & Uses Morphological \bzd prompting and NFD-normalization. & \textbf{19.99} \\
\bzd & 80 & 20 & Uses Regular Prompting. & 11.50 \\
\bzd & 40 & 20 & ... & 11.41 \\
\bzd & 10 & 10 & ... & 11.41 \\
\bzd & 20 & 20 & ... & 11.16 \\
\bzd & 20 & 10 & ... & 10.95 \\
\bzd & 40 & 10 & ... & 10.95 \\
\bzd & 10 & 10 & ... & 10.63 \\
\bzd & 80 & 10 & ... & 10.17 \\
\bzd & 80 & 0  & ... & 6.53 \\
\bzd & 40 & 0  & ... & 5.93 \\
\bzd & 20 & 0  & ... & 5.30 \\
\bzd & 10 & 0  & ... & 4.75 \\

\midrule
\multicolumn{5}{c}{\textbf{\textit{\grn}}} \\
\midrule
Language & \multicolumn{1}{c}{Retrieval (r)} & \multicolumn{1}{r}{Dev Exemplars (d)} & \multicolumn{1}{|l|}{Notes} & \multicolumn{1}{r}{chrF++} \\
\midrule
\grn & 40 & 49 & Includes synthetic exemplars. Not submitted because ablation was incomplete at submission deadline. & 51.34 \\
\grn & 40 & 20 & ... & 48.38 \\
\grn & 80 & 49 & Includes synthetic exemplars. & \textbf{48.24} \\
\grn & 80 & 20 & ... & 42.61 \\
\grn & 0  & 49 & ... & 20.80 \\

\midrule
\multicolumn{5}{c}{\textbf{\textit{\nlv}}} \\
\midrule
Language & \multicolumn{1}{c}{Retrieval (r)} & \multicolumn{1}{r}{Dev Exemplars (d)} & \multicolumn{1}{|l|}{Notes} & \multicolumn{1}{r}{chrF++} \\
\midrule
\nlv & 40 & 20 & -- & \textbf{25.67} \\
\nlv & 40 & 10 & -- & 25.59 \\
\nlv & 80 & 20 & -- & 25.25 \\
\nlv & 80 & 10 & -- & 25.16 \\
\nlv & 20 & 20 & -- & 24.16 \\
\nlv & 20 & 10 & -- & 23.91 \\
\nlv & 40 & 0  & -- & 16.56 \\
\nlv & 80 & 0  & -- & 16.37 \\
\nlv & 20 & 0  & -- & 15.61 \\

\midrule
\multicolumn{5}{c}{\textbf{\textit{\hch}}} \\
\midrule
Language & \multicolumn{1}{c}{Retrieval (r)} & \multicolumn{1}{r}{Dev Exemplars (d)} & \multicolumn{1}{|l|}{Notes} & \multicolumn{1}{r}{chrF++} \\
\midrule
\hch & 40 & 20 & -- & \textbf{18.99} \\
\hch & 40 & 10 & -- & 17.74 \\
\hch & 20 & 20 & -- & 17.56 \\
\hch & 80 & 10 & -- & 17.48 \\
\hch & 80 & 20 & -- & 17.13 \\
\hch & 20 & 10 & -- & 16.81 \\
\hch & 40 & 0  & -- & 16.59 \\
\hch & 80 & 0  & -- & 16.25 \\
\hch & 20 & 0  & -- & 13.80 \\

\midrule
\multicolumn{5}{c}{\textbf{\textit{\yua}}} \\
\midrule
Language & \multicolumn{1}{c}{Retrieval (r)} & \multicolumn{1}{r}{Dev Exemplars (d)} & \multicolumn{1}{|l|}{Notes} & \multicolumn{1}{r}{chrF++} \\
\midrule
\yua & -- & 49 & Fixes $r = 0$ for lack of retrieval exemplars. & \textbf{26.29} \\
\yua & -- & 20 & ... & 26.29 \\
\yua & -- & 40 & ... & 25.07 \\
\yua & -- & 30 & ... & 25.05 \\
\yua & -- & 0 & Fixes $r = 0$ for lack of retrieval exemplars. At $d = 0$, the model receives no signal from the target language. & 20.25 \\

\bottomrule
\end{tabular}
\caption{Dev chrF++ results across languages for the proposed system. (...) Indicates previous \textit{Notes} column entry applies. (--) Indicates no \textit{Notes}. \textbf{Bold} entries \textit{ChrF++} column are submission scores for each target language.}
\label{tab:grid_search}
\end{table*}


\end{document}